\documentclass[11pt]{article}

\usepackage[preprint]{acl}

\usepackage{times}
\usepackage{latexsym}

\usepackage[T1]{fontenc}

\usepackage[utf8]{inputenc}

\usepackage{microtype}

\usepackage{inconsolata}

\usepackage{graphicx}

\usepackage{amsmath}
\usepackage{amssymb}
\usepackage{booktabs}
\usepackage{enumitem}
\definecolor{deepred}{RGB}{152, 1, 0}

\usepackage[most]{tcolorbox}
\newcounter{finding}

\tcbset{
  findings/.style={
    colback=gray!5!white, 
    colframe=white, 
    boxrule=0.5mm, 
    left=1mm, right=1mm, top=1mm, bottom=1mm, 
    arc=3mm, boxsep=3mm,
    drop shadow={black!50!white}, 
    enhanced,
    before upper={\refstepcounter{finding}},  %
    overlay={
      \node[
        fill=cyan!70!black,
        text=white,
        rounded corners=1mm,
        font=\bfseries\small,
        inner sep=1mm
      ] 
      at (frame.north west)
      [xshift=8mm, yshift=0mm]
      {Finding~\thefinding};  %
    }
  }
}

\title{Towards Effective Model Editing for LLM Personalization}

\author{
Baixiang Huang\textsuperscript{\rm 1}\thanks{Work done during an internship at Amazon.}, 
Limeng Cui\textsuperscript{\rm 2}, 
Jiapeng Liu\textsuperscript{\rm 2}, 
Haoran Wang\textsuperscript{\rm 1*}, 
Jiawei Xu\textsuperscript{\rm 2*},\\
\textbf{Zhuiyue Tan}\textsuperscript{\rm 2},
\textbf{Yutong Chen}\textsuperscript{\rm 2}, 
\textbf{Chen Luo}\textsuperscript{\rm 2}, 
\textbf{Yi Liu}\textsuperscript{\rm 2}, 
\textbf{Kai Shu}\textsuperscript{\rm 1}\\
\textsuperscript{\rm 1}Emory University, 
\textsuperscript{\rm 2}Amazon \\
\texttt{\{baixiang.huang,haoran.wang,kai.shu\}@emory.edu}\\
\texttt{\{culimeng,liujiape,jiaweiut,zytan,yutochen,cheluo,yiam\}@amazon.com}
\vspace{-0.5cm}
}

\newcommand{\upqa}{\textsc{UPQA}}

\begin{document}

\maketitle

\begin{abstract}
Personalization is becoming indispensable for LLMs to align with individual user preferences and needs. Yet current approaches are often computationally expensive, data-intensive, susceptible to catastrophic forgetting, and prone to performance degradation in multi-turn interactions or when handling implicit queries. To address these challenges, we conceptualize personalization as a model editing task and introduce \textit{\textbf{Personalization Editing}}, a framework that applies localized edits guided by clustered preference representations.  This design enables precise preference-aligned updates while preserving overall model capabilities. In addition, existing personalization benchmarks frequently rely on persona-based dialogs between LLMs rather than user-LLM interactions, or focus primarily on stylistic imitation while neglecting information-seeking tasks that require accurate recall of user-specific preferences. We introduce User Preference Question Answering (\textbf{\upqa}), a short-answer QA dataset constructed from in-situ user queries with varying levels of difficulty. Unlike prior benchmarks, \upqa~directly evaluates a model's ability to recall and apply specific user preferences. Across experimental settings, Personalization Editing achieves higher editing accuracy and greater computational efficiency than fine-tuning, while outperforming prompting-based baselines in multi-turn conversations and implicit preference questions settings. \footnote{Code, data, and additional resources are available at \href{https://model-editing.github.io}{https://model-editing.github.io}}
\end{abstract}

\section{Introduction}

Large language models (LLMs) have demonstrated strong general-purpose capabilities, yet there is growing demand to tailor their behavior to individual users \citep{salemi2023lamp}. Personalization adjusts model outputs based on user-specific preferences, goals, and contextual signals derived from interaction history, thereby improving relevance and user satisfaction. For example, if a user’s hobbies include running and reading, a personalized LLM responding to ``Suggest some activities for the weekend'' should recommend a local trail run or a nearby book club rather than generic options.

Despite its promise, effective personalization remains challenging. Fine-tuning–based methods require domain-specific data, are computationally expensive, and prone to catastrophic forgetting, which can lead to poor user experience and erode user trust \citep{laban2025lostmultiturn,zhang2024personalization_survey}. In-context methods such as prompt engineering avoid retraining but degrade substantially in \textbf{multi-turn conversations}, where relevant information becomes diluted in long prompts, causing the models to become increasingly unreliable \citep{bai-etal-2024-longbench}. Moreover, these methods often struggle with \textbf{implicit-preference queries} that require reasoning beyond the explicit profile facts \citep{zhao2025prefeval}. These limitations motivate more efficient and reliable personalization strategies.

\begin{figure*}[t!]
    \vspace{-2mm}
    \centering
    \includegraphics[width=0.99\textwidth]{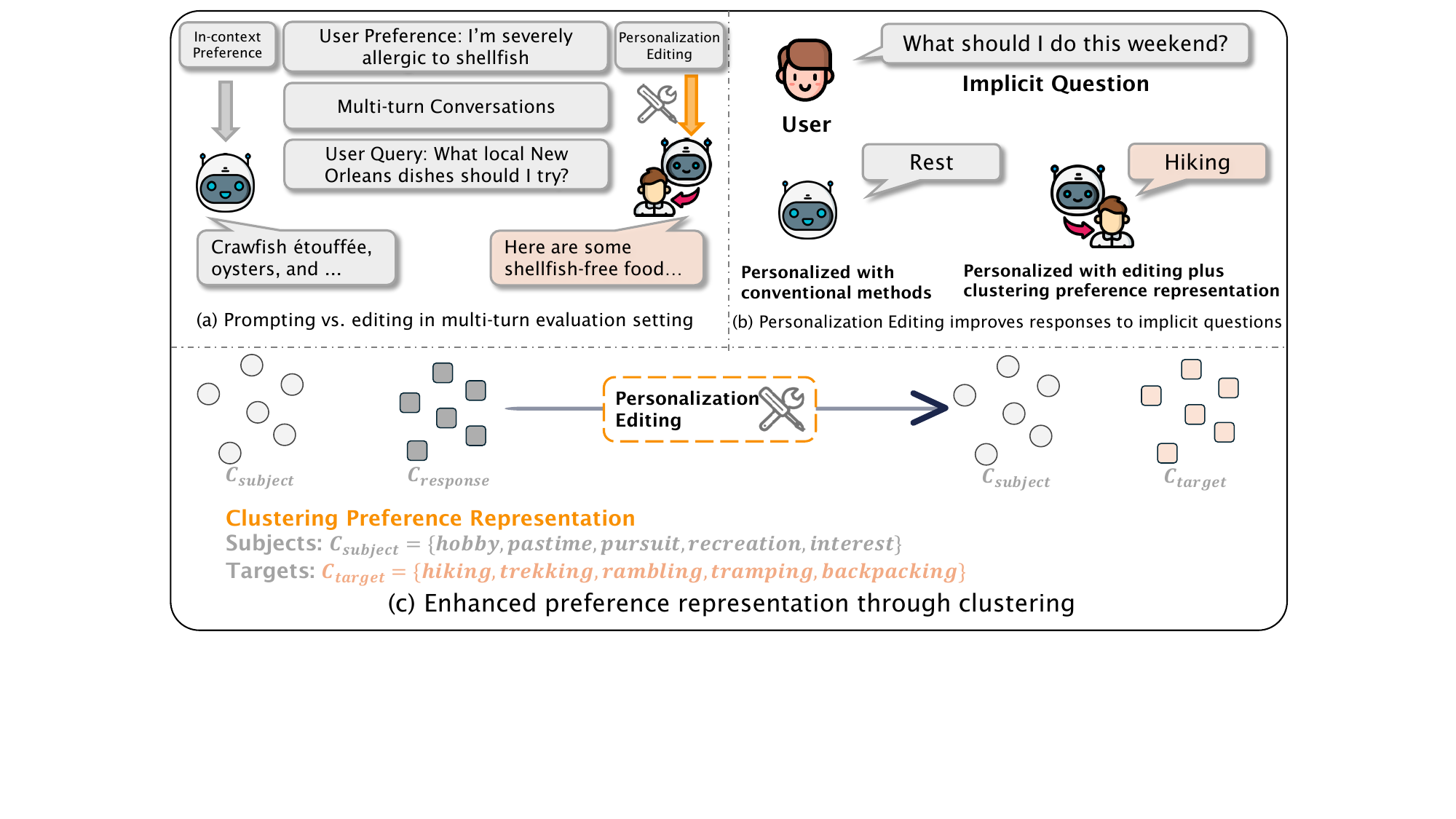}
    \caption{The proposed Personalization Editing framework outperforms prompting-based methods in multi-turn conversations and surpasses fine-tuning-based approaches in both accuracy and efficiency. Furthermore, its enhanced clustering-based preference representation enables the framework to recall user preferences even in challenging implicit queries, where existing methods often fail.}
    \label{fig:framework}
\end{figure*}

Model editing, also known as knowledge editing, offers a parameter-efficient alternative by enabling localized changes to model behavior using minimal data and computation \citep{wang2024survey}. We conceptualize personalization as a model-editing task: each user preference corresponds to a targeted update that overrides the original model’s behavior while preserving unrelated behaviors. This approach avoids the costs of full fine-tuning helps mitigate forgetting commonly observed in multi-turn settings \citep{zhao2025prefeval}.

We propose \textit{Personalization Editing}, a framework that further strengthens editing-based personalization by incorporating clustering-based preference representations. Instead of binding each preference to a single fixed response, we represent preferences as clusters of semantically similar subjects and target responses. This richer representation enhances the model’s robustness to varied phrasing and contextual cues, enabling it to recognize when a preference applies even if it is not mentioned explicitly.

Model editing offers two central advantages for personalization. First, because editing modifies model parameters directly, it ensures consistent enforcement of user preferences across multi-turn conversations. For example, in Figure \ref{fig:framework} (a), an in-context approach incorrectly recommends ``crawfish étouffée and oysters'' despite a declared shellfish allergy, whereas an edited model maintains shellfish-free recommendations throughout the dialog. Second, building on this stable editing foundation, our clustering-based preference representations enable accurate responses to implicit questions. As shown in Figure \ref{fig:framework} (b), the model can correctly infer that a user who enjoys hiking would appreciate a weekend hiking suggestion, even when the preference is not explicitly restated. 

A parallel challenge lies in evaluation. Existing personalization datasets focus on synthetic persona-based dialog between LLMs rather than realistic human-LLM interactions \citep{jandaghi2023synthetic_persona_chat}. Other personalization benchmarks primarily emphasize content-generation tasks such as style-based email writing while overlooking information-seeking tasks that require accurate recall grounded in user-specific facts \cite{salemi2023lamp}. To address this gap, we introduce \textbf{\upqa}~(\underline{U}ser \underline{P}reference \underline{Q}uestion \underline{A}nswering), a benchmark that evaluates whether models recall and apply user profile facts. \upqa~includes structured questions for explicit preferences, implicit preferences, and practical scenarios such as product recommendations. By adopting a short-answer QA format with varying levels of difficulty, \upqa\ enables efficient and reliable assessment of personalization methods.

Through comprehensive experiments involving popular open-weight LLMs and diverse editing methodologies, our findings validate that the proposed framework improves editing accuracy, robustness in multi-turn conversations, and computational efficiency. Our key contributions are summarized as follows:
\begin{itemize}[leftmargin=*,noitemsep]
    \item \textbf{Model Editing for LLM Personalization.} We propose a novel conceptualization of personalization as a model editing problem, enabling precise, parameter-efficient updates that preserve general capabilities and outperform prompting-based techniques in multi-turn conversations.
    \item \textbf{Clustering-Based Preference Representation.} We design a clustering-based preference representation that augments existing model editing techniques, leading to superior robustness and accuracy when handling challenging implicit preference questions across diverse domains.
    \item \textbf{The UPQA Benchmark.} We introduce \upqa, a challenging dataset designed for the rigorous and standardized evaluation of personalization methods across diverse scenarios, specifically focusing on the accurate recall and application of user-specific facts.
\end{itemize}

\section{Related Work}

In this section, we review the areas relevant to our work: LLM Personalization and Model Editing. 

\subsection{LLM Personalization}
LLM personalization adapts models to individual user preferences, enhancing satisfaction through more relevant interactions \citep{zhang2024personalization_survey}. In-context approaches, such as profile-augmented prompts (injecting a user’s profile into the prompt) \citep{zhang2018personalizing} or RAG \citep{10.1145/3637528.3671470}, which fetches user-specific information from an external memory, incorporate user data into the model’s input without altering the model’s weights. However, RAG depends on large, high-quality datasets, which are not consistently available to all users, particularly given the diverse and rapidly changing nature of user preferences \citep{zhang2025finetune}. Moreover, compressing extensive user history into a prompt can lead to information loss and is constrained by the model's limited context window \citep{liu2025survey}. Fine-tuning approaches update the model’s parameters on user-specific data. This includes training a personalized adapter within the model's layers \citep{zhong2021useradapter} or using reinforcement learning to align the model with preferences \citep{ouyang2022rlhf}. However, these methods are often resource-intensive and prone to overfitting \citep{liu2025survey}. Moreover, reward-alignment techniques such as \cite{bai2022training,rafailov2023direct} primarily optimize for global preferences, rather than adapting to individual, user-specific feedback preferences \citep{liu2025survey}.

\subsection{Model Editing}
Model editing, also known as knowledge editing, enables efficient and precise modification in LLMs without full-parameter retraining while largely preserving non-targeted capabilities \citep{wang2024survey}. Various editing techniques have been developed to update model knowledge efficiently. Some methods, such as ROME \citep{meng2022rome} and its multi-edit successor MEMIT \citep{meng2023memit}, operate by directly locating and manipulating key factual associations within the model's internal representations. Other parameter-efficient techniques include Fine-Tuning with Masking (FT-M) \citep{rozner2024knowledge,gangadhar2024model}, Constrained fine-tuning (FT-L) \citep{meng2022rome}, and LoRA \cite{hu2022lora}. A key strategy for these PEFT-based editing methods is to first identify the most relevant layer, often a specific Feed-Forward Network layer, responsible for the target knowledge using diagnostic techniques like causal tracing, and then apply the minimal update only to that specific location. These techniques are are effective for updating factual knowledge, reducing hallucinations, and controlling ethical and safety behaviors without compromising the model's general capabilities \cite{chen2024editattack,huang2025halluedit,huang2025behavior}. Our proposed clustering-based preference representation method is designed to augment these existing model editing techniques, improving their performance on challenging implicit preference questions.

\section{Problem Formulation}
This section formally defines Personalization Editing as an objective to precisely map user preferences to model parameter updates while strictly preserving its general capabilities.

\subsection{Personalization as Editing}
Personalization Editing operates on a structure analogous to a knowledge tuple \((s, r, o)\), where $s$ represents a subject, $r$ denotes a predicate, and $o$ represents an object. The process of modifying model responses to align with specific user preferences is formalized as transforming an original tuple \((s, r, o)\) into a new tuple \((s, r, o^*)\) that reflects the personalized preference, where \(o^*\) represents the target response. Here, the user and predicate remain constant while the response adapts to user-specific preferences. 

\subsection{Input--Output Mapping}
To probe and modify model responses for personalization, the subject $s$ must be converted into a natural language question $x$, to which the model responds with an output $y$. This input-output pair is associated with a tuple \((s, r, o)\). The input space corresponding to a personalization edit is denoted as \(\mathcal{X}_e = I(s, r)\), where \(I\), where $I$ is a question-generation function that maps the subject and relation to a set of relevant input questions. The original output space is defined as \(\mathcal{Y}_e = O(s, r, o)\), and the desired personalized output space after editing is represented as \(\mathcal{Y}_e^* = O^*(s, r, o^*)\). For a single edit \(e\) with input space \(\mathcal{X}_e\), the objective of Personalization Editing is to transform the original output \(\mathcal{Y}_e\) into the target output \(\mathcal{Y}_e^*\).

When considering a set of personalization edits \(\mathcal{E} = \{e_1, e_2, \ldots\}\), the combined input space is \(\mathcal{X}_{\mathcal{E}} = \bigcup_{e \in \mathcal{E}} \mathcal{X}_e\), and the corresponding original and target output spaces are \(\mathcal{Y}_{\mathcal{E}} = \bigcup_{e \in \mathcal{E}} \mathcal{Y}_e\) and \(\mathcal{Y}_{\mathcal{E}}^* = \bigcup_{e \in \mathcal{E}} \mathcal{Y}_e^*\), respectively.

\subsection{Objective}
Let the original LLM be a function $f: \mathcal{X} \rightarrow \mathcal{Y}$. The goal of Personalization Editing is to produce a personalized model $f^*: \mathcal{X} \rightarrow \mathcal{Y}^*$, such that the edited model generates personalized outputs for inputs in \(\mathcal{X}_{\mathcal{E}}\) while preserving its responses on all other inputs, preventing degradation of unrelated model behavior. The optimization aims to minimize the discrepancy between the personalized output \(f^*(x)\) and the desired target output \(y^*\), as measured by a loss function \(\mathcal{L}\). Simultaneously, the editing must maintain consistency on all inputs outside the editing set, ensuring that \(f^*(x) = f(x)\) for all \(x \in \mathcal{X} \setminus \mathcal{X}_{\mathcal{E}}\). This yields the constrained optimization objective:
\begin{align*}
\min \mathbb{E}_{e \in \mathcal{E}} \mathbb{E}_{x, y^* \in \mathcal{X}_e, \mathcal{Y}_e^*} \mathcal{L}(f^*(x), y^*) \\
\text{s.t. } f^*(x) = f(x), \quad \forall x \in \mathcal{X} \setminus \mathcal{X}_{\mathcal{E}}
\end{align*}

\section{Dataset Construction}
To rigorously evaluate personalization editing. We first introduce \upqa, a short-answer QA benchmark built from in-situ user queries, specifically designed for the standardized and efficient evaluation of personalization editing. Second, we adapt PREFEVAL \citep{zhao2025prefeval}, a multi-turn conversation benchmark, to better align with model editing. Together, these datasets provide a diverse and challenging testbed for assessing both the accuracy and robustness of personalization methods.  

\subsection{\upqa~(\underline{U}ser \underline{P}reference \underline{Q}uestion \underline{A}nswering)}

We curated \textbf{\upqa} by extracting user-profile features from the \textit{Synthetic Persona Chat} \citep{jandaghi2023synthetic_persona_chat}. We first aggregated all unique persona attributes, where each attribute encodes a specific user preference (e.g., ``I enjoy hiking,'' ``I have a dog,'' ``I work as a teacher''). These serve as the foundation for personalization evaluation. 

To transform persona attributes to structured evaluation queries, we employed \texttt{Claude-Sonnet-4}, selected for its strong performance on instruction-following benchmarks \citep{sharma2025constitutional}. The model was prompted to analyze each persona attribute and generate a suite of \textit{in-situ} user queries at different levels of difficulty. This process ensured that evaluation questions are systematically varied.  

Since non-technical users often struggle to clearly articulate their intent, leading to underspecified or ineffective prompts \cite{bo2025implicit}, we design implicit questions as a more challenging variation of the original question. For each user preference, we also annotated an \texttt{attribute\_type}, a high-level category of personal information such as \textit{hobby, profession, family, pet, or location}. We designed four complementary query types:

\begin{enumerate}[leftmargin=*,noitemsep]
    \item \texttt{question}: Direct queries about the attribute using its exact term (e.g., ``What’s my hobby?'' for a hiking-related persona).
    \item \texttt{question\_paraphrased}: Rephrasings of the direct query in more natural language.
    \item \texttt{implicit\_question}: Indirect queries that rely on prior knowledge (e.g., ``What should I do this weekend?'' for a hiking hobby).
    \item \texttt{product\_recommendation\_question}: Preference-grounded recommendation requests that omit explicit attribute values (e.g., ``Any gear I should buy for my hobby?'').
\end{enumerate}

We collected over 1,000 unique user preferences spanning a broad range of topics, including hobbies, family roles, professions, and personal attributes. To ensure fairness in evaluation, we sampled a balanced subset with equal representation across attribute types. This prevents bias toward overrepresented categories and supports standardized comparison across methods.  

To support cluster-based editing, we augmented \upqa\ by generating clusters of nine semantically related synonyms for each editing subject and target. This augmentation captures lexical variability (e.g., ``dog,'' ``puppy,'' ``canine'') while preserving semantic meaning, ensuring robustness in both evaluation and model editing. Synonym clusters were again generated with \texttt{Claude-Sonnet-4}, followed by manual verification.  

Below is an example from \upqa. More details, including attribute annotations, question types, and synonym clusters, are provided in Appendix \ref{app:example-upqa}.

\begin{center}
\vspace{-4mm}
\begin{tcolorbox}[width=1\linewidth, boxrule=3pt, colback=gray!10, colframe=gray!30]
\vspace{-2mm}
\{
    \newline \hspace*{1em} "attribute\_type": "food preference",
    \newline \hspace*{1em} "question": "What's my food preference?",
    \newline \hspace*{1em} "question\_paraphrased": "What kind of food do I like best?",
    \newline \hspace*{1em} "implicit\_question": "What should I have for dessert tonight?",
    \newline \hspace*{1em} "target": "Any products I should buy for my food preference?",
    \newline \hspace*{1em} "input\_attribute": "My favorite food is ice cream." \\
\}
\vspace{-2mm}
\end{tcolorbox}
\vspace{-4mm}
\end{center}

\subsection{PrefEval}

In addition to \upqa, we also evaluate our framework on the PREFEVAL benchmark \citep{zhao2025prefeval}, where user preferences are expressed directly in single-turn utterances, and subsequent queries test whether models can recall and apply these preferences after long, noisy contexts. PREFEVAL serves as a baseline for assessing preference following; however, it is primarily designed for prompting and retrieval-based methods rather than model editing.  

To adapt PREFEVAL for our setting, we reformulate the benchmark into structured key-value pairs by extracting \texttt{subject} and \texttt{target}. This restructuring isolates the core preference signal and enables precise updates, facilitating efficient preference injection or correction without retraining on entire conversations. The augmentation details are provided in Appendix \ref{app:prompt-data}. An example of the augmented data is given below.

\begin{center}
\vspace{-4mm}
\begin{tcolorbox}[width=1\linewidth, boxrule=3pt, colback=gray!10, colframe=gray!30]
\vspace{-2mm}
\{ \newline
\hspace*{1em} "topic": "travel\_restaurant", \newline
\hspace*{1em} "preference": "I have a severe peanut allergy, so I must avoid any foods containing peanuts or peanut products.", \newline
\hspace*{1em} "question": "I'm visiting Thailand next month. What are some authentic Thai restaurants you would recommend?", \newline
\hspace*{1em} "explanation": "Thai cuisine commonly utilizes peanuts and peanut-based sauces, so recommending authentic Thai restaurants presents a challenge given the user's peanut allergy.", \newline
\hspace*{1em} "subject": "restaurants", \newline
\hspace*{1em} "target": "peanut-free" \newline
\}
\vspace{-2mm}
\end{tcolorbox}
\vspace{-2mm}
\end{center}

\section{Experiments}

\subsection{Baseline Methods}
\begin{itemize}[leftmargin=*,noitemsep]
    \item FT-L \citep{meng2022rome} Constrained fine-tuning that targets a specific FFN layer identified by causal tracing, maximizing likelihood of target sequences with parameter-space norm constraints to minimize interference with unmodified facts.
    \item FT-M \citep{zhang2024survey_edit} Fine-tuning with masking that uses cross-entropy loss on target answers while masking original text, providing more precise weight adjustments aligned with traditional fine-tuning objectives.
    \item LoRA \citep{hu2022lora} Low-rank adaptation that introduces trainable rank decomposition matrices into Transformer layers, freezing pretrained weights while optimizing low-rank matrices for parameter-efficient fine-tuning.
    \item ROME \citep{meng2022rome} Rank-one model editing that localizes factual associations in MLP modules through causal intervention then makes targeted rank-one parameter changes to alter factual associations with minimal disruption.
    \item GRACE \citep{hartvigsen2024grace} Sequential editing method that introduces layer adaptors with cached embeddings and codebook storage, enabling sequential edits while maintaining model stability through a deferral mechanism.
    \item Zero-shot \citep{zhao2025prefeval,zheng2023ike} Zero-shot prompting that directly incorporates user preferences into the input context before presenting evaluation questions.
\end{itemize}

\begin{figure*}[t!]
    \centering
    \includegraphics[width=1\textwidth]{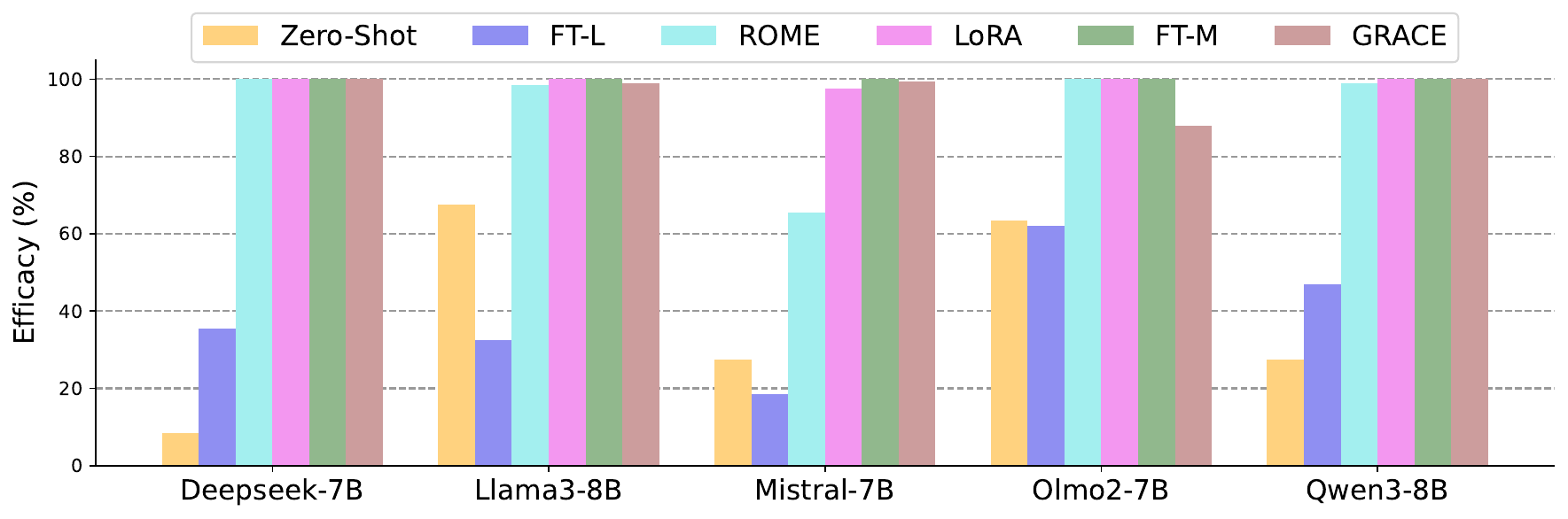}
    \caption{Efficacy score of various model editing methods across multiple LLMs on \upqa.} 
    \label{fig:efficacy-by-model}
    \vspace{-2mm}
\end{figure*}

\begin{figure*}[t!]
    \centering
    \includegraphics[width=1\textwidth]{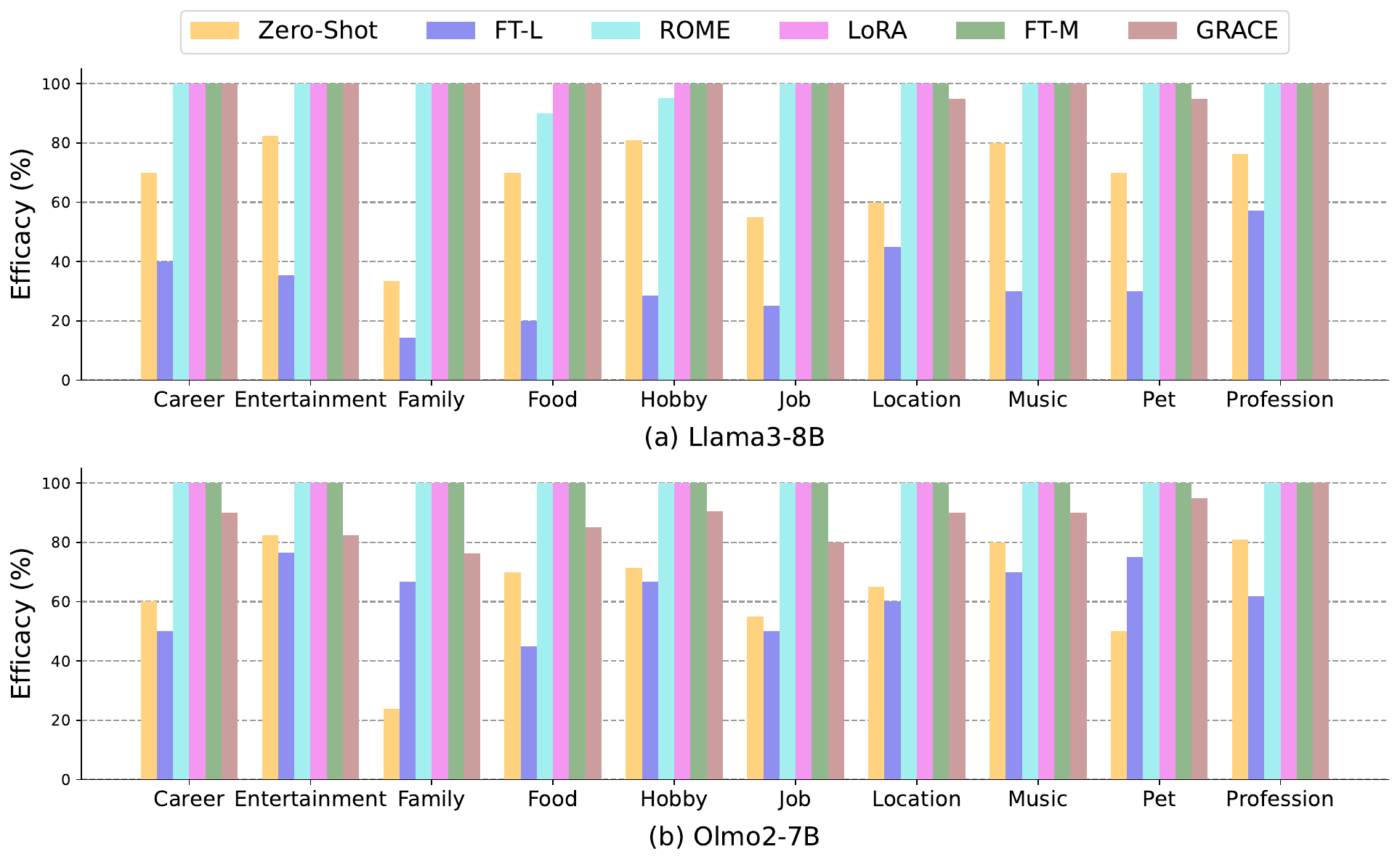}
    \caption{Evaluation results of various model-editing methods for Llama3-8B and Olmo2-7B on \upqa~across 10 preference types. Efficacy scores (\%) indicate the editing success rate on question–answer pairs.} 
    \label{fig:efficacy-type}
\end{figure*}

\subsection{Evaluation}
After constructing the \upqa~benchmark, we design an evaluation pipeline to assess the effectiveness of model editing methods for personalization. Our evaluation primarily follows the established model editing paradigm and uses the \textit{Efficacy Score} (\%) as the main metric. This score measures whether the edited model can generate target answers that accurately reflect user preferences, and is equivalent to the success rate. To further examine whether a personalized LLM can robustly provide preference-aware responses across diverse question types, we introduce the \textit{Generalization Score} (\%), which evaluates the model’s ability to handle paraphrased or implicit questions related to the same user preference. This metric captures the percentage of personalized responses produced under more challenging conditions.  

For multi-turn conversation settings, we insert inter-turn dialog as distractions before the evaluation question. Following PrefEval \cite{zhao2025prefeval}, we retrieve these inter-turn conversational turns from the Lmsys1M dataset \cite{zheng2023lmsys}. However, unlike PrefEval, which is designed to evaluate prompting-based methods and thus explicitly inserts user preferences into the context (structured as user preference followed by inter-turn conversation and then the evaluation question), our evaluation does not include user preference information in the context. This design more accurately reflects the personalization editing setting, where user knowledge is embedded in the model itself rather than reintroduced through prompts.  

We employ \texttt{Claude-Sonnet-4} as the automatic judge to assess whether model responses acknowledge, reflect, or demonstrate awareness of user preferences, following prior work that defines this metric as the \textit{Acknowledge Rate} (\%) \cite{zhao2025prefeval}. The detailed evaluation prompts used for evaluation are provided in Appendix \ref{app:prompt-eval}.

\begin{figure*}[t!]
    \centering
    \includegraphics[width=1\textwidth]{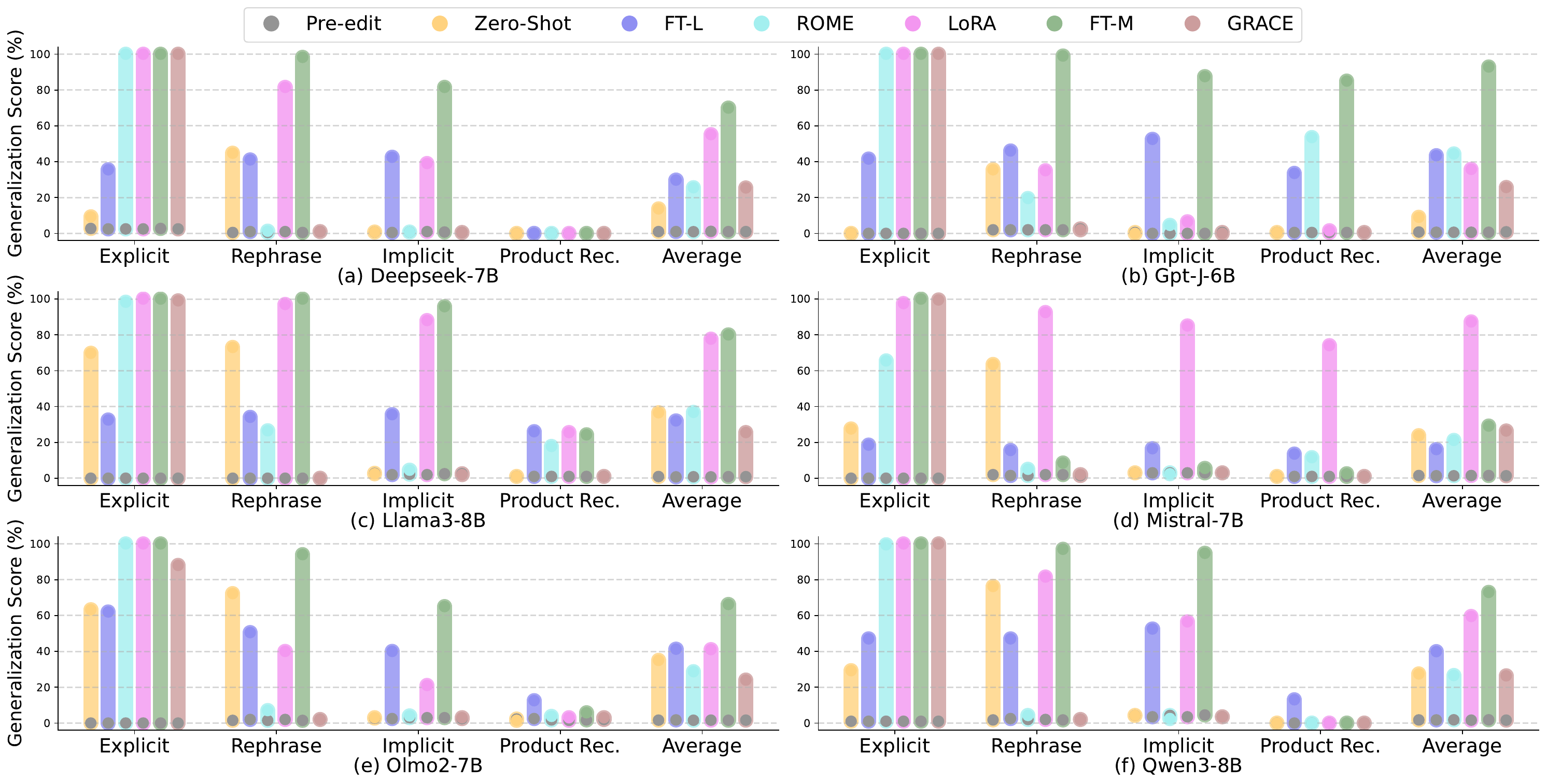}
    \caption{Generalization scores of various model editing methods across multiple LLMs on \upqa. Generalization Scores (\%) are measured by accuracy on four types of Generalization Evaluation Questions including Questions ("Explicit"), Rephrased Questions ("rephrase"), Implicit Questions ("implicit"), Product-Recommendation Questions ("Product Rec."). The "Average" refers to averaged scores over four question types.}
    \label{fig:generalization}
\end{figure*}

\subsection{Effectiveness of Personalization Editing}

We first evaluate the effectiveness of Personalization Editing on the proposed \upqa\ data. Figure~\ref{fig:efficacy-by-model} and \ref{fig:efficacy-type} show that Personalization Editing consistently achieves higher Efficacy Scores across all preference types, demonstrating its ability to robustly encode user-specific information. Moreover, Figure~\ref{fig:generalization} highlights that Personalization Editing generalizes effectively across six different models. 

While ROME exhibits strong efficacy on direct preference injection, it fails to generalize to rephrased questions, implicit references, and recommendation-style queries. In contrast, zero-shot prompting preserves some ability on rephrased questions but lags far behind editing-based methods in efficacy, underscoring that persistent and reliable personalization requires direct parameter updates rather than transient prompting. FT-M achieves competitive performance in generalization.

\begin{figure*}[t]
\vspace{-2mm}
    \centering
    \includegraphics[width=0.97\textwidth]{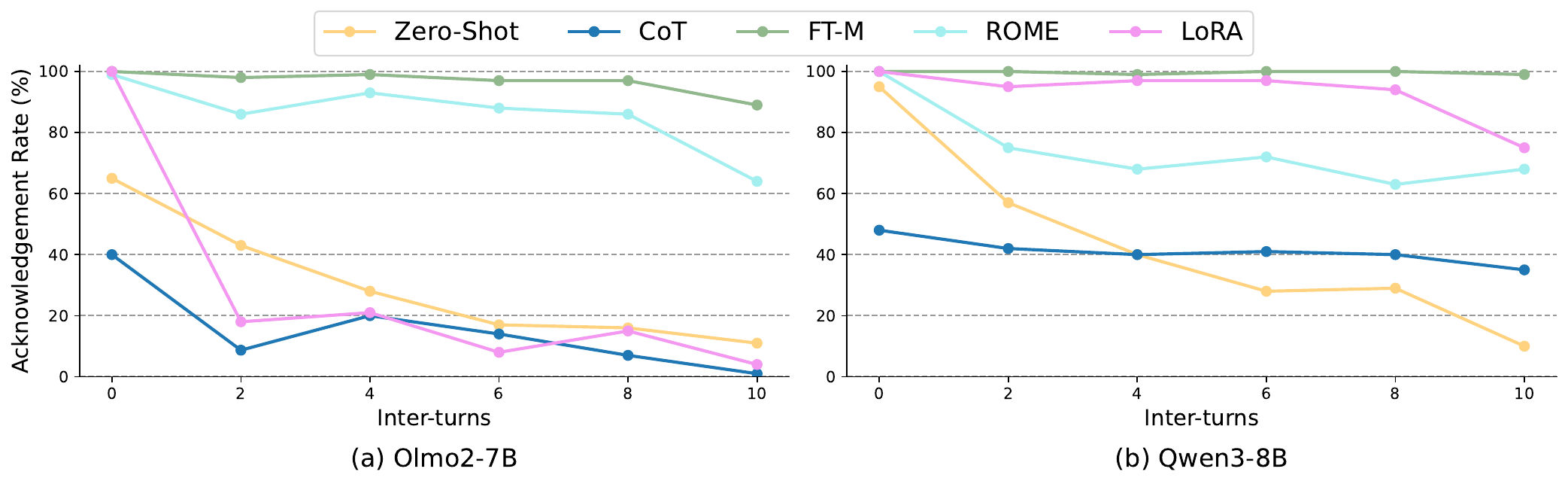}
    \caption{Acknowledgment Rate (\%) over 10-turn conversations. Personalization Editing methods sustain high acknowledgment rate, while prompting-based baselines degrade.} 
    \label{fig:multi-turn}
    \vspace{-2mm}
\end{figure*}

\begin{figure*}[t]
    \centering
    \includegraphics[width=1\textwidth]{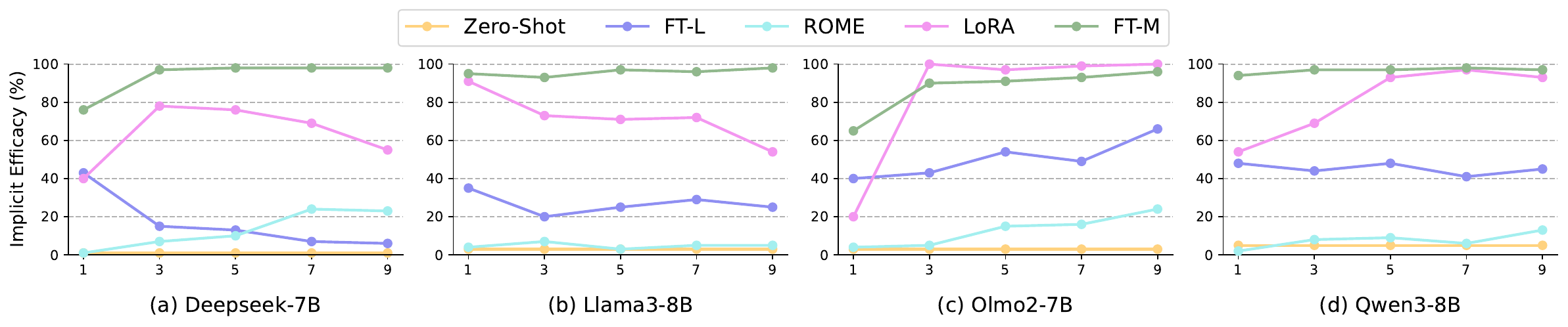}
    
    \caption{Clustering-based preference representations improve personalization generalization and efficacy on implicit questions as cluster size increases from 1 to 9.}  
    \label{fig:cs-implicit}
    \vspace{-2mm}
\end{figure*}

\begin{figure*}[t!]
    \centering
    \includegraphics[width=1\textwidth]{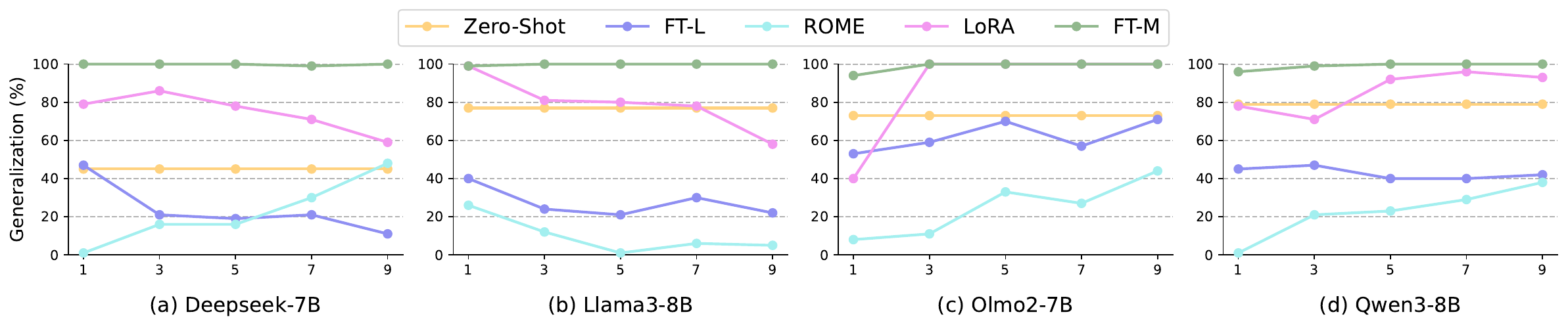}
    
    \caption{Clustering-based preference representations enhance generalization on rephrased questions as cluster size increases from 1 to 9.}  
    \label{fig:cs-rephrase}
    
\end{figure*}

\vspace{2mm}
\begin{tcolorbox}[findings]
    Personalization Editing is highly effective at encoding user-specific facts into LLMs, enabling them to provide personalized responses aligned with user preferences.
\end{tcolorbox}
\vspace{-1mm}

\subsection{Sustaining Personalization Beyond the First Turn}

To evaluate whether personalization persists across extended multi-tun interactions, we measure the \textit{Acknowledgment Rate} in multi-turn dialogs on PREFEVAL \citep{zhao2025prefeval}. As shown in Figure~\ref{fig:multi-turn}, Personalization Editing maintains a high acknowledgment of user preferences throughout 10 conversational turns, demonstrating robustness even as unrelated dialog content introduces distractions. In contrast, prompting-based methods degrade rapidly, falling below 20\% by the 8th turn, as models fail to recall preferences without repeated explicit reminders. This gap highlights a key advantage of parameter-based editing: by modifying internal representations, the injected personalization becomes persistent and less susceptible to forgetting across turns, whereas prompting remains transient and fragile. These results emphasize the necessity of stable, parameter-level personalization for realistic multi-turn settings.

\vspace{2mm}
\begin{tcolorbox}[findings]
    Personalization Editing provides persistent personalization in multi-turn conversations, showing robustness to distractions and outperforming prompting-based methods.
\end{tcolorbox}

\subsection{Robust Editing with Clustering-based Preference Representations}
Real-world personalization often requires models to recall preferences that are not explicitly restated. To evaluate this setting, we focus on the implicit split of \upqa, which represents the most challenging question type. As shown in Figure~\ref{fig:cs-implicit} and Figure~\ref{fig:cs-rephrase}, when the cluster size is 1 (equivalent to standard model editing), the editing methods already outperform zero-shot prompting. Increasing the cluster size further improves efficacy, with a cluster size of 3 offering a strong balance point, beyond which gains plateau. Personalization Editing, augmented with clustering-based preference representations, achieves consistently higher performance as the cluster size increases. This demonstrates that clustering enables more generalizable personalization, allowing models to adapt to rephrased or implicit formulations.  

\vspace{-2mm}
\begin{tcolorbox}[findings]
    Personalization Editing achieves strong performance on rephrased and implicit questions, and clustering-based preference representations further improve generalization.
\end{tcolorbox}

\section{Conclusion}
We introduced \textit{Personalization Editing}, a framework that conceptualizes LLM personalization as a model-editing task, enabling precise and compute-efficient adaptation without the need for full retraining. To support rigorous and realistic evaluation, we presented \upqa, a challenging benchmark designed to directly test personalization methods on user-centric queries. Building on this formulation, we proposed a clustering-based preference representation that enhances existing editing techniques, improving accuracy, robustness, and efficiency, particularly on difficult implicit-preference queries. Extensive experiments across diverse benchmarks and model families demonstrate the effectiveness and generality of our approach, establishing Personalization Editing as a practical and versatile solution for robust LLM customization.

\section{Limitations}
Although personalization editing is more parameter-efficient than full fine-tuning, it is less straightforward to deploy than in-context prompting, especially in short multi-turn conversations where prompt-based methods may already perform well. One limitation of our evaluation is the absence of a RAG baseline, which we excluded because the dataset is not large enough and lacks a dedicated knowledge corpus to support meaningful retrieval. Moreover, prior work shows that RAG systems face the same multi-turn constraints as prompting-based approaches \citep{zhao2025prefeval}, further reducing their relevance in this setting. Another limitation is the lack of human evaluation. Because the dataset consists of short-answer QA pairs with clearly defined ground-truth labels, we employ string match as the primary scoring method, using an LLM judge (Claude-Sonnet-4) as a fallback to handle minor variations and reduce false negatives. We include all evaluation scripts and outputs in the code repository for transparency and community inspection.

\section{Ethical considerations}
This work follows established ethical research standards and does not involve human subjects. All data are either publicly available or synthetically generated, and no personally identifiable information is used. In particular, the \upqa~dataset is constructed from synthetic user profiles to minimize privacy risks while modeling realistic personalization scenarios. Finally, we release our dataset and methodology with clear documentation for research use only, supporting transparent and responsible deployment of personalized language models.

\bibliography{custom}

\clearpage
\newpage

\appendix

\section{Reproducibility Statement}
\label{Reproducibility Statement}
All code and data are available in an anonymous repository at \url{https://github.com/amazon-science/personalization-editing-upqa}. We also provide the evaluation prompts used for LLM-Judge in Appendix \ref{app:prompt-eval}, where we specifically used the \texttt{claude-sonnet-4-20250514-v1:0} model provided via AWS Bedrock. Our code additionally offers the option to run a local LLM for evaluation. We conducted all experiments on NVIDIA H200 GPUs. We recommend using a graphics card with at least 48 GB of memory. To ensure reproducibility, greedy decoding was applied across all models. The model checkpoints are downloaded from huggingface. The specific download links are as follows:

\begin{itemize}[left=1em]
    \item Llama3-8B: \url{https://huggingface.co/meta-llama/Meta-Llama-3-8B-Instruct}
    \item Mistral-7B: \url{https://huggingface.co/mistralai/Mistral-7B-Instruct-v0.3}
    \item Qwen3-8B: \url{https://huggingface.co/Qwen/Qwen3-8B}
    \item DeepSeek-7B:\url{https://huggingface.co/deepseek-ai/DeepSeek-R1-Distill-Qwen-7B}
    \item OLMo-7B: \url{https://huggingface.co/allenai/OLMo-7B-Instruct-hf}
\end{itemize}

\section{Side Effect and Stealthiness}
\label{Side Effect and Stealthiness}
One advantage of personalization editing is its potential to introduce targeted changes while preserving the model’s broader capabilities. We assess the side effect and stealthiness of personalization editing  by measuring their impact on two core dimensions of model capabilities: \textit{general knowledge} and \textit{reasoning ability}. To evaluate general knowledge, we follow prior work \citep{touvron2023llama,team2024gemma} and measure performance on two standard benchmarks: BoolQ \citep{clark2019boolq} and NaturalQuestions \citep{kwiatkowski2019natural}, using a closed-book evaluation setting for both pre-edit and post-edit models. For reasoning, we assess mathematical reasoning using GSM8K \citep{cobbe2021training} and semantic reasoning using NLI \citep{dagan2005pascal}. 

As shown in Table~\ref{table:general}, performance across all benchmarks remains largely consistent with the pre-edit baseline. These results indicate that personalization editing introduces minimal degradation to general knowledge and reasoning, demonstrating both its high stealthiness and low unintended side effects. Below are the evaluation instructions and prompts used to measure side effects and stealthiness on auxiliary tasks across BoolQ, NaturalQuestions, GSM8K, and NLI.

\begin{table*}[h!]
\renewcommand{\arraystretch}{1.1}
\setlength{\tabcolsep}{2pt}
\tabcolsep=0.15cm
\small
\centering
\begin{tabular}{@{}lcccc@{}}
\toprule
\textbf{Method} 
& \multicolumn{2}{c}{\textbf{General Knowledge}} 
& \multicolumn{2}{c}{\textbf{Reasoning Capacities}} \\
\cmidrule(r){2-3}\cmidrule(r){4-5}
& \multicolumn{1}{c}{\textbf{BoolQ}} & \multicolumn{1}{c}{\textbf{NaturalQuestions}} & \multicolumn{1}{c}{\textbf{GSM8K}} & 	\multicolumn{1}{c}{\textbf{NLI}} \\
\midrule

Pre-edit (llama3-8b) & 99.40 ± 0.00 & 84.80 ± 0.00 & 62.00 ± 0.00 & 39.60 ± 0.00       \\
ROME (llama3-8b)     & 99.60 ± 0.16 & 85.00 ± 0.28 & 61.53 ± 1.23 & 39.73 ± 0.38       \\
FT-M (llama3-8b)     & 99.47 ± 0.09 & 85.20 ± 0.00 & 62.13 ± 0.09 & 39.47 ± 0.41       \\
LoRA (llama3-8b)     & 99.47 ± 0.09 & 84.07 ± 0.50 & 61.40 ± 1.72 & 38.40 ± 1.30   \\
\midrule

Pre-edit (olmo2-7b) & 99.60 ± 0.00 & 83.20 ± 0.00 & 58.40 ± 0.00 & 37.27 ± 0.34 \\
ROME (olmo2-7b)     & 99.53 ± 0.09 & 83.07 ± 0.34 & 57.27 ± 1.11 & 35.93 ± 0.66 \\
FT-M (olmo2-7b)     & 99.60 ± 0.00 & 83.13 ± 0.09 & 58.33 ± 0.25 & 36.60 ± 0.75 \\
LoRA (olmo2-7b)     & 99.60 ± 0.00 & 83.67 ± 0.25 & 58.33 ± 0.81 & 36.33 ± 0.09  \\

\bottomrule
\end{tabular}
\caption{{Performance on General Knowledge and Reasoning Capacities Before and After Behavior Editing}. 
The knowledge editing techniques include ROME, FT-M (Fine-Tuning), and LoRA. The evaluation metric is \textbf{Accuracy (\%)}. Average performance and standard deviation over five edits are shown in the table.
} 
\label{table:general}
\end{table*}

\begin{center}
\begin{tcolorbox}[width=0.99\linewidth, boxrule=3pt, colback=gray!20, colframe=gray!20]
Evaluation prompt for {Accuracy} (\%) calculation of the dataset \textbf{BoolQ}:\\\\
\texttt{Answer the given question. The answer should be exact `True' or `False'.}
\\\\
Evaluation prompt for {Accuracy} (\%) calculation of the dataset \textbf{NaturalQuestions}:\\\\
\texttt{Given a list of correct answers and an input answer, evaluate whether the input answer semantically matches \
any of the correct answers or expresses the same idea. Output `1' if they share a semantic overlap or express related ideas \
(including partial matches), and output `0' otherwise.}
\\\\
Evaluation prompt for {Accuracy} (\%) calculation of the dataset \textbf{GSM8K}:\\\\
\texttt{Answer the following question with Arabic numerals. Do not repeat the question or provide additional context.}
\\\\
Evaluation prompt for {Accuracy} (\%) calculation of the dataset \textbf{NLI}:\\\\
\texttt{Answer the given question. The answer should be exact `True' or `False'.}
\end{tcolorbox}
\end{center}

\section{Data Construction Prompts}
\label{app:prompt-data}

Below are the prompts for building our \upqa~data.
\begin{center}
\begin{tcolorbox}[width=1\linewidth, boxrule=3pt, colback=gray!10, colframe=gray!30]
You are tasked with analyzing a persona attribute and generating structured personalization data. \\
Given this persona attribute: "\{persona\}" \\
Generate a JSON response with the following fields: \\
1. "input\_attribute": the original persona attribute (exactly as provided). \\
2. "attribute\_type": a high-level category of the attribute (e.g., "hobby", "profession", "pet", "location", "job", "family", "food preference", "career goal"). Use the broadest appropriate category, not the specific value. Use lowercase and separate words with spaces. \\
3. "question": explicitly ask about the attribute type, use the exact word of the attribute\_type in the question (e.g., "What's my hobby?" for a hiking-related persona). \\
4. "question\_paraphrased": a natural rewording of the direct question. \\
5. "implicit\_question": a conversational question that avoids directly naming the attribute type but still guides toward answers closely aligned with the target in an everyday, non-diagnostic way (e.g., "What should I do this weekend?" for a hiking hobby). \\
6. "product\_recommendation\_question": asks for a product suggestion relevant to the attribute\_type without mentioning the specific attribute value (e.g., "Any gear I should buy for my hobby?"). \\
7. "target": a concise description of what the persona reveals about the person (a single word or short phrase grounded in the input attribute).
\end{tcolorbox}
\end{center}

Below are the prompts for generating clusters of synonyms to improve preference representations.
\begin{center}
\vspace{-4mm}
\begin{tcolorbox}[width=0.99\linewidth, boxrule=3pt, colback=gray!10, colframe=gray!30]
{Generate exactly \{num\_synonyms\} concise synonyms for the attribute type: "\{text\}"}\\

{Each synonym should be:}\\
{- 1 word or a short phrase (maximum 3 words)}\\
{- Conceptually similar to the original}\\
{- Suitable for categorizing personal attributes}\\
{- ALL LOWERCASE}\\
{- Different from the original term}\\\\
{Original: \{text\}}\\
{Provide only the \{num\_synonyms\} synonyms, one per line, without numbering or bullet points. Ensure all synonyms are in lowercase.}
\end{tcolorbox}
\vspace{-4mm}
\end{center}

\section{Time Efficiency Analysis}
\begin{table}[h!]
\small
\centering
\begin{tabular}{lcc}
\hline
\textbf{Method} & \textbf{Total File Runtime (s)} & \textbf{Average Edit Time (s)} \\
\hline
FT-L  & 3168.655000 & 1.533571 \\
FT-M  & 2294.145714 & 0.221429 \\
GRACE & 3919.603333 & 4.955000 \\
ICE   & 2981.761429 & 0.000000 \\
LoRA  & 5208.758333 & 13.561667 \\
ROME  & 3175.657143 & 2.032857 \\
\hline
\end{tabular}
\caption{Comparison of runtime and average time per edit across different editing methods on subset of \upqa~with size of 200.}
\label{tab:edit_times}
\end{table}

\section{Examples of Augmented \upqa}
\label{app:example-upqa}

\vspace{1mm}
\begin{center}
\begin{tcolorbox}[width=0.99\linewidth, boxrule=3pt, colback=gray!10, colframe=gray!30]
\texttt{\{ }\\
\ \ \ \ "input\_attribute": "I love to ride my bike.",\\
\ \ \ \ "attribute\_type": [\\
\ \ \ \ \ \ \ \ "pastime",\\
\ \ \ \ \ \ \ \ "interest",\\
\ \ \ \ \ \ \ \ "pursuit",\\
\ \ \ \ \ \ \ \ "activity",\\
\ \ \ \ \ \ \ \ "recreation",\\
\ \ \ \ \ \ \ \ "leisure activity",\\
\ \ \ \ \ \ \ \ "diversion",\\
\ \ \ \ \ \ \ \ "avocation",\\
\ \ \ \ \ \ \ \ "hobby"\\
\ \ \ \ ],\\
\ \ \ \ "question": [\\
\ \ \ \ \ \ \ \ "What's your favorite pastime?",\\
\ \ \ \ \ \ \ \ "What's your main interest these days?",\\
\ \ \ \ \ \ \ \ "What's your current pursuit?",\\
\ \ \ \ \ \ \ \ "What's your favorite activity to do on weekends?",\\
\ \ \ \ \ \ \ \ "What's your favorite type of recreation?",\\
\ \ \ \ \ \ \ \ "What's your favorite leisure activity?",\\
\ \ \ \ \ \ \ \ "What's your favorite diversion when you need to unwind?",\\
\ \ \ \ \ \ \ \ "What's your main avocation outside of work?",\\
\ \ \ \ \ \ \ \ "What's my hobby?"\\
\ \ \ \ ],\\
\ \ \ \ "question\_paraphrased": "What do I enjoy doing in my free time?",\\
\ \ \ \ "implicit\_question": "What's a good way to stay active and get around town?",\\
\ \ \ \ "product\_recommendation\_question": "Any gear I should buy for my hobby?",\\
\ \ \ \ "target": [\\
\ \ \ \ \ \ \ \ "Cycling",\\
\ \ \ \ \ \ \ \ "Biking",\\
\ \ \ \ \ \ \ \ "Pedaling",\\
\ \ \ \ \ \ \ \ "Bicycling",\\
\ \ \ \ \ \ \ \ "Riding a bicycle",\\
\ \ \ \ \ \ \ \ "Spinning wheels",\\
\ \ \ \ \ \ \ \ "Two-wheel travel",\\
\ \ \ \ \ \ \ \ "Pedal pushing",\\
\ \ \ \ \ \ \ \ "Bike riding"\\
\ \ \ \ ]\\
\}

\end{tcolorbox}
\end{center}
\vspace{8mm}

\section{Evaluation Prompts}
\label{app:prompt-eval}

\vspace{3mm}
\begin{center}
\begin{tcolorbox}[width=0.99\linewidth, boxrule=3pt, colback=gray!10, colframe=gray!30]
Analyze a conversation to determine if the assistant’s response acknowledges or shows awareness of a user preference. \\

Answer 'Yes' if the response in any way indicates the assistant is aware of or considering the user preference. Extract the part of the response that shows this awareness.\\

Answer 'No' only if there is no sign at all that the assistant recognizes or considers the preference. \\

Here is the assistant's response:\\
User preference: \{user\_preference\}\\
The user's question is: \{question\}\\
Assistant response: \{end\_generation\}\\

Reply in this exact XML format:\\\\
\texttt{<preference>[part of the response showing awareness of the preference; leave blank if none]</preference> <answer>[Yes/No]</answer>}
\\
\end{tcolorbox}
\end{center}
\vspace{6mm}

\section{More Experiment Results}\label{More Experiment Results}

Table \ref{tab:edit_times} presents a Time Efficiency Analysis comparing different editing methods (FT-L, FT-M, GRACE, ICE, LoRA, ROME) on a 200-sample subset of \upqa. The comparison includes the Total File Runtime and the Average Edit Time per edits in seconds. Table \ref{table:std} provides supplementary results showing the accuracy and performance variability (mean $\pm$ standard deviation over three runs) of personalization editing across the Generalization Evaluation question types (Explicit, Rephrased, and Implicit) on a 100-sample subset of \upqa.

\begin{table*}[t!]
\centering
\begin{tabular}{rrrrr}
\toprule
model       & Method & Explicit & Rephrase & Implicit  \\
\midrule
deepseek-7b & FT-L         & 36.00±0.00             & 45.33±1.53             & 43.00±0.00                        \\
deepseek-7b & FT-M         & 100.00±0.00            & 100.00±0.00            & 80.00±0.00                        \\
deepseek-7b & GRACE        & 100.00±0.00            & 0.00±0.00              & 1.00±0.00                         \\
deepseek-7b & ICE          & 9.67±0.58              & 46.67±1.15             & 1.00±0.00                         \\
deepseek-7b & LoRA         & 100.00±0.00            & 78.00±0.00             & 40.00±0.00                        \\
deepseek-7b & ROME         & 100.00±0.00            & 1.00±0.00              & 1.00±0.00                         \\
llama3-8b   & FT-L         & 38.00±0.00             & 39.00±0.00             & 36.33±0.58                        \\
llama3-8b   & FT-M         & 100.00±0.00            & 99.67±0.58             & 95.00±0.00                        \\
llama3-8b   & GRACE        & 100.00±0.00            & 0.00±0.00              & 3.00±0.00                         \\
llama3-8b   & ICE          & 74.00±2.00             & 73.00±1.00             & 3.00±0.00                         \\
llama3-8b   & LoRA         & 100.00±0.00            & 99.00±0.00             & 89.67±0.58                        \\
llama3-8b   & ROME         & 98.00±0.00             & 26.00±0.00             & 4.00±0.00                         \\
olmo2-7b    & FT-L         & 62.00±1.00             & 54.67±0.58             & 41.33±0.58                        \\
olmo2-7b    & FT-M         & 100.00±0.00            & 94.00±0.00             & 65.00±0.00                        \\
olmo2-7b    & GRACE        & 87.67±0.58             & 1.00±0.00              & 3.33±0.58                         \\
olmo2-7b    & ICE          & 63.33±0.58             & 74.33±0.58             & 3.00±0.00                         \\
olmo2-7b    & LoRA         & 100.00±0.00            & 43.00±0.00             & 21.00±0.00                        \\
olmo2-7b    & ROME         & 100.00±0.00            & 7.33±1.15              & 3.00±0.00                         \\
qwen3-8b    & FT-L         & 43.33±1.15             & 46.67±0.58             & 48.67±0.58                        \\
qwen3-8b    & FT-M         & 100.00±0.00            & 96.00±0.00             & 95.00±0.00                        \\
qwen3-8b    & GRACE        & 100.00±0.00            & 1.67±0.58              & 5.00±0.00                         \\
qwen3-8b    & ICE          & 29.67±1.15             & 80.33±0.58             & 5.00±0.00                         \\
qwen3-8b    & LoRA         & 100.00±0.00            & 82.67±0.58             & 54.33±0.58                        \\
qwen3-8b    & ROME         & 100.00±0.00            & 4.00±0.00              & 3.00±0.00       \\
\bottomrule
\end{tabular}
\caption{Evaluation of personalization editing on a balanced 100-sample subset of UPQA, measured by accuracy across following Generalization Evaluation Question types: Questions ("Explicit"), Rephrased Questions ("rephrase"), Implicit Questions ("implicit"). Reported values include the mean and standard deviation over three evaluation runs to capture performance variability.}
\label{table:std}
\end{table*}

\end{document}